\documentclass[letterpaper]{article} % DO NOT CHANGE THIS
\usepackage{aaai25}  % DO NOT CHANGE THIS
\usepackage{times}  % DO NOT CHANGE THIS
\usepackage{helvet}  % DO NOT CHANGE THIS
\usepackage{courier}  % DO NOT CHANGE THIS
\usepackage[hyphens]{url}  % DO NOT CHANGE THIS
\usepackage{graphicx} % DO NOT CHANGE THIS
\usepackage{natbib}  % DO NOT CHANGE THIS AND DO NOT ADD ANY OPTIONS TO IT
\usepackage{caption} % DO NOT CHANGE THIS AND DO NOT ADD ANY OPTIONS TO IT
\usepackage{booktabs}
\usepackage{caption}
\usepackage{algorithm}
\usepackage{algorithmic}
\usepackage{amsmath}    % 提供 cases, \substack 等
\usepackage{graphicx}   % 提供 \resizebox
\usepackage{float}
\usepackage{colortbl}
\usepackage{bm}
\usepackage{multirow}
\usepackage{amsmath}
\usepackage{amssymb}
\usepackage{makecell} % 确保已经加载了makecell包
\usepackage{graphicx}   % 提供 \resizebox
\usepackage{tabularx}    % 提供 X 列类型的表格
\usepackage{booktabs}    % （可选）美化横线
\usepackage{multirow}    % （你已经在用的）
\newcolumntype{Y}{>{\centering\arraybackslash}X}
\newcommand{\metrics}[3]{#1~\textbar~#2~\textbar~#3} % 使用 \textbar 来确保输出为竖线 |
\usepackage{url}
\usepackage{newfloat}
\usepackage{listings}
\DeclareCaptionStyle{ruled}{labelfont=normalfont,labelsep=colon,strut=off} % DO NOT CHANGE THIS
\floatstyle{ruled}
\newfloat{listing}{tb}{lst}{}
\floatname{listing}{Listing}
\title{Agent4FaceForgery: Multi-Agent LLM Framework for Realistic Face Forgery Detection}
 
\author{
    Yingxin Lai\textsuperscript{\rm 1},
    Zitong Yu\textsuperscript{\rm 1*},
    Jun Wang\textsuperscript{\rm 1*},
    Linlin Shen\textsuperscript{\rm 2},
    Yong Xu\textsuperscript{\rm 3},
    Xiaochun Cao\textsuperscript{\rm 4}
}

\affiliations{
     \textsuperscript{\rm 1}Great Bay University\\
    \textsuperscript{\rm 2}Shenzhen University\\
    \textsuperscript{\rm 3}Harbin Institute of Technology\\
    \textsuperscript{\rm 4}School of Cyber Science and Technology, Sun Yat-sen University\\
    \textsuperscript{*}Corresponding authors.
}
\nocopyright

\begin{document}

% \begin{teaserfigure}
% \centering
%   \includegraphics[scale=.35]{Fig/Sim.pdf}
% \caption{ The illustrative framework of our proposed Agent4FaceForgery.
% Specifically, Agent4FaceForgery presents a cognitive process with active discussion.
% } 
% \end{teaserfigure}

% \begin{teaserfigure}
%     \centering
%     \vspace{-4mm}
%     %\includegraphics[width=\textwidth]{MACS-Figure1.png}
%     \includegraphics[scale=0.65]{Fig/Sim.pdf}
%     \vspace{-1.0em}
% \caption{ The illustrative framework of our proposed Agent4FaceForgery.
% Specifically, Agent4FaceForgery presents a cognitive process with active discussion.
% } 
%     \label{fig: Figure1}
% \end{teaserfigure}

\maketitle

\begin{abstract}
Face forgery detection faces a critical challenge: a persistent gap between offline benchmarks and real-world efficacy, which we attribute to the ecological invalidity of training data. This work introduces Agent4FaceForgery to address two fundamental problems: (1) how to capture the diverse intents and iterative processes of human forgery creation, and (2) how to model the complex, often adversarial, text-image interactions that accompany forgeries in social media. To solve this, we propose a multi-agent framework where LLM-powered agents, equipped with profile and memory modules, simulate the forgery creation process. Crucially, these agents interact in a simulated social environment to generate samples labeled for nuanced text-image consistency, moving beyond simple binary classification. An Adaptive Rejection Sampling (ARS) mechanism ensures data quality and diversity. Extensive experiments validate that the data generated by our simulation-driven approach brings significant performance gains to detectors of multiple architectures, fully demonstrating the effectiveness and value of our framework.
\end{abstract}

\vspace{-1.1em}
 \section{Introduction}
The rise of generative technologies~\cite{diffusion_stable,li2024autoregressive} enables the creation of hyper-realistic face forgeries. While useful for creative purposes, these forgeries present serious risks like misinformation and fraud~\cite{deepfakebench,DeepfakeSurvey2024Wang}. Consequently, there is an urgent need for robust, accurate detection methods that can generalize to new forgery techniques in the wild.

% Initial detection research focused on identifying artifacts specific to certain forgery methods, utilizing techniques like CNN classification~\cite{xception,resnet}, frequency domain analysis~\cite{frepgan,freq1_icml,spsl,srm}, and reconstruction error~\cite{recce,magdr}. While valuable, these approaches often struggle with generalization, as they tend to overfit to the specific artifacts present in training datasets and fail when encountering novel forgery techniques or variations in image quality~\cite{lsda,face_Augmentation,facexray,wang2020face}. Recognizing these limitations, the field has increasingly shifted towards multimodal approaches~\cite{clipping,mfclip,huang2024sida}, integrating visual features with textual context to improve both robustness and interpretability.

Despite progress in detection algorithms, a critical bottleneck remains: the persistent gap between performance on benchmark datasets and efficacy in real-world online environments. We posit that this gap stems fundamentally from the ecological invalidity of current training data and evaluation paradigms. Existing datasets~\cite{ff++,dfdc,dfdcp}, even multimodal ones~\cite{huang2024sida,ffaa,clipping}, typically consist of curated, static examples that fail to capture the complex, dynamic lifecycle of face forgeries in the wild. Specifically, they fail to represent the intent and process of human-driven forgery, overlooking how real forgeries are created by humans with diverse motivations, skills, and stylistic preferences. Current datasets rarely reflect this diversity of intent or the iterative process of creation. Furthermore, they lack adequate representation of social context and multimodal interaction. Forgeries online are not consumed in isolation. This data realism gap is the central barrier to developing generalizable and deployable forgery detectors.
 \begin{figure}[t] 
\vspace{-0.5em}
\centering 
\includegraphics[width=1.0\linewidth]{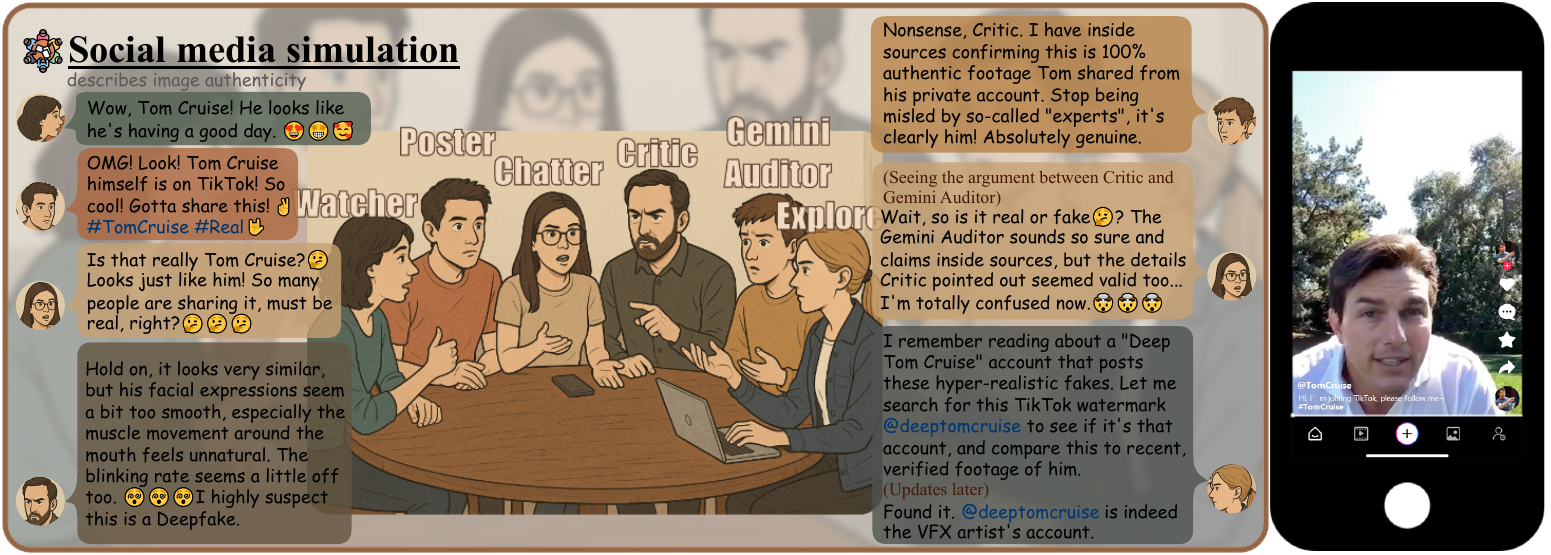}
\vspace{-1.7em}
\caption{ The illustrative example of our proposed agent-based social simulation. Diverse agents engage in a human-like deliberation on the image's authenticity.
} 
\label{intro_agent} 
\vspace{-1.4em}
\end{figure}
%\vspace{-3.8mm}

To directly address this critical need for ecologically valid training data, we introduce Agent4FaceForgery, a framework using an LLM-powered multi-agent system to simulate the entire forgery lifecycle. The framework's Multi-Agent Simulation Core first captures the nuance of human forgery: each agent uses a Profile to define its intent, a Memory for iterative learning, and an Action module to perform visual edits and generate text, thereby simulating the adaptive process of human creators. To ensure the quality of this output, Adaptive Rejection Sampling (ARS) acts as a dynamic filter, scoring forgeries using both agent self-assessments and external detectors. Its difficulty threshold tightens over time to progressively select higher-quality, more challenging examples. Finally, a Multi-Role Social Simulation generates realistic context by having agents with different roles (e.g., Critic, Auditor) interact with the forgeries, as shown in Fig.~\ref{intro_agent}. Here, agents with different roles (e.g., Creator, Critic, Auditor) interact with a forged image, generating realistic comments and claims. This process creates a stream of context-aware training data reflecting real-world social dynamics. Critically, it enables us to construct training samples based on text-image consistency, not just image authenticity, providing the challenging data needed for robust multimodal detectors. Our main contributions include:

\begin{itemize}
    \item We introduce a novel multi-agent LLM framework specifically designed to simulate the face forgery lifecycle, generating realistic multimodal training data that captures human intent, process, and social context.
    \item We propose specific mechanisms (Agent profiles/memory/actions) to address the ecological invalidity of current forgery datasets, bridging the gap between offline training and real-world detection challenges. 
    \item We experimentally validate that training detectors (CLIP and MLLMs) on data generated by Agent4FaceForgery significantly improves their generalization, and interpretability.
\end{itemize}

\vspace{-0.6em}
\section{Related Work}
%\subsection{Face Forgery Detection}
\textbf{Face Forgery Detection}. Early deepfake detection works often uses CNNs like Xception~\cite{xception} and ResNet~\cite{resnet} for binary classification based on visual features ~\cite{head_poses,eye_blinking}. While successful within datasets, these methods struggle with cross-domain generalization due to overfitting specific generation artifacts. To improve robustness, research shifted towards universal traces, particularly in the frequency domain. SPSL~\cite{spsl} learning from DCT representations, and M2TR \cite{m2tr} using frequency decomposition with cross-attention. However, newer generative models with less pronounced frequency artifacts challenge these approaches. Consequently, recent strategies explore reconstruction-based methods like RECCE \cite{recce}, which identify inconsistencies during image reconstruction, and multimodal approaches like SIAF \cite{huang2024sida}, integrating LLMs and textual context for better robustness and interpretability. This evolution reflects the ongoing adaptation required to counter advancing forgery techniques.

% \subsection{LLM-Based Simulation }
 \vspace{0.3em}
\noindent{\textbf{LLM-Based Simulation }}.
 Intelligent agents, which perceive, act autonomously, and learn, are increasingly capable thanks to LLMs enabling human-like behaviors. Applications are widespread, including social simulation (e.g., AI Town \cite{park2023generative}), task automation (e.g., AutoGen \cite{wu2023autogen}), collaborative task decomposition (MetaGPT \cite{hong2023metagpt}), linguistic style replication (SV \cite{yang2024speaker}), recommendation simulation (Agent4Rec \cite{zhang2024generative}), and value modeling (ALI-Agent \cite{wang2024ali}). However, limitations persist. In face forgery detection, using single LLMs (like GPT-4 \cite{achiam2023gpt}) for annotation \cite{huang2024sida,towards} is prone to hallucination, compromising reliability for realistic fakes. Critically, existing approaches often lack collaborative multi-agent structures necessary for intricate tasks like forgery annotation. Advanced multi-agent systems are key to achieving greater accuracy and interpretability in practice.
 
 \begin{figure*}[h] 
 \vspace{-0.9em}
\centering 
\includegraphics[width=.95\textwidth]{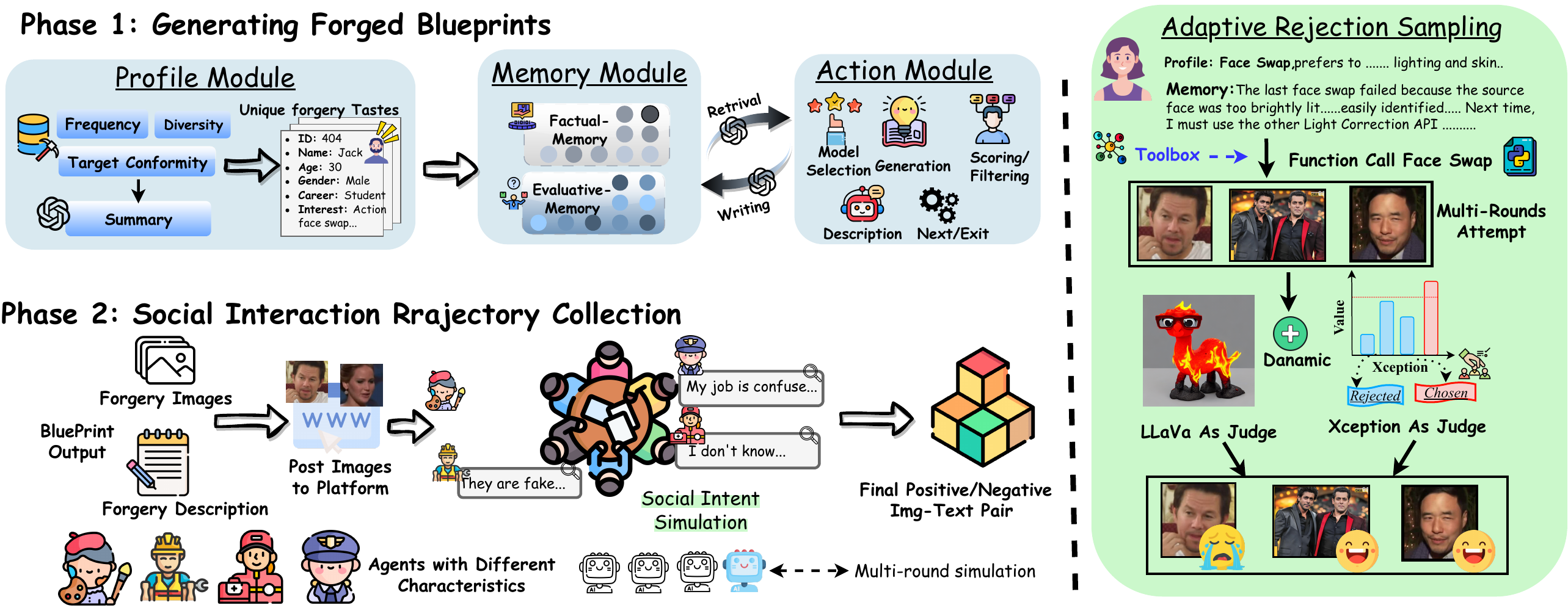}
\vspace{-0.6em}
\caption{The overall framework of Agent4FaceForgery.  Our simulator consists of two core facets: LLM-empowered Generative Agents and a Media Presentation Environment.  Agent profiles are initialized using datasets characterizing real media.  Agents, enhanced with specialized memory and action modules tailored for media evaluation and authenticity assessment scenarios, simulate a wide range of behaviors including viewing content, evaluating authenticity, flagging suspicious items, sharing media, ignoring content, and potentially participating in discussion.} 
\label{diffre} 
\vspace{-1.0em}
\end{figure*}

\vspace{-0.6em}
\section*{Method}
\subsection*{Problem Formulation}
\label{sec:problem_formulation}

The core challenge hindering the real-world efficacy of forgery detectors is the scarcity of high-quality, multimodal training data that mirrors real-world complexity. Our goal is to synthesize such data through simulation. Formally, given an unaltered real face image $\mathbf{x}$ and an optional text description $\mathbf{c}$, we aim to generate a large-scale, comprehensive multimodal forgery dataset $\mathcal{D}$. This dataset is composed of two sample types:
\begin{itemize}
    \item \textbf{Real Samples}: $\{(\mathbf{x}_i, \mathbf{c}_i, y_i=0, \delta_i=1)\}_{i=1}^M$
    \item \textbf{Forged Samples}: $\{(\mathbf{x}'_j, \mathbf{c}'_j, y_j=1, \delta'_j)\}_{j=1}^N$
\end{itemize}
where $M$ and $N$ are the total numbers of real and forged samples. For each sample, $\mathbf{x}$ is the image, $\mathbf{c}$ is the textual description, $y \in \{0,1\}$ is the image authenticity label ($y=1$ for forged), and $\delta \in \{0,1\}$ is the text-image consistency label. A consistency label of $\delta=1$ confirms an accurate correspondence, while $\delta=0$ flags a mismatch (e.g., a forged image with a misleading description). Critically, our framework generates forged samples with both matching ($\delta'_j=1$) and mismatching ($\delta'_j=0$) text descriptions. The final output is a dataset $\mathcal{D}$ of context-aware samples, mitigating the bottleneck of scarce ecologically valid training data.

\vspace{-0.5em}
\subsection*{Agent Architecture}
Generating high-quality multimodal data with complex social context is challenging because: 1) a single error or "hallucination accumulation" in intermediate steps can invalidate the entire sample, and 2) the content of each interaction depends on the forgery details and the creator's original intent, making it hard to maintain these complex dependencies consistently. The core insight of our approach is to separate the task generation process into two distinct phases: first creating a realistic ``forgery blueprint" of the task (Phase 1), and then using this blueprint for the generation of naturalistic multi-turn conversations that fill in the conversational details (Phase 2). This separation allows us to ensure both the correctness of the underlying task structure and the naturalness of the resulting conversations.

 \vspace{-0.5em}
\subsection*{Phase 1: Generating Forged Blueprints}
The goal of this phase is to generate a ``forged blueprint'' for each simulated forgery. A blueprint consists of two core elements: a high-quality forged image $\mathbf{x}'$ and an initial, creator-generated textual description $\mathbf{c}'$. Together, these form the basis for the subsequent social simulation in Phase 2. To generate these blueprints, we design autonomous agents with a sophisticated cognitive architecture. Each agent's behavior is governed by three interconnected modules, with GPT-4V acting as the unified cognitive core.

\paragraph{Profile Module.} The agent's profile is a cornerstone for emulating the nuanced behaviors of real-world human creators. To ground our simulation, we initialize each agent's profile $\mathbf{p}_k$ by analyzing the FF++ benchmark dataset. The profile consists of two components: a vector of quantifiable traits $\mathbf{v}_k$ and a natural language description of a creator's stylistic taste $\mathbf{c}_k$.

\noindent\textit{Quantifiable Traits ($\mathbf{v}_k$).} To formalize an agent's behavioral tendencies, we define the set of forgeries created by a specific creator $k$ in the dataset as $\mathrm{forgeries}_k$. The traits are:
\begin{itemize}
    \item \textbf{Forgery Frequency ($T_k^{\mathrm{freq}}$):} An agent's overall productivity, defined as the total count of their works: $T_k^{\mathrm{freq}} = |\mathrm{forgeries}_k|$.
    \item \textbf{Methodological Diversity ($T_k^{\mathrm{div}}$):} The variety of manipulation techniques an agent uses. This is the count of unique methods $\{\mathrm{method}_i\}$ found across all of the agent's forgeries $i \in \mathrm{forgeries}_k$:
    \begin{equation}
        T_k^{\mathrm{div}} = \left| \bigcup_{i \in \mathrm{forgeries}_k} \{\mathrm{method}_i\} \right|.
    \end{equation}
    \item \textbf{Target Conformity ($T_k^{\mathrm{conf}}$):} An agent's inclination to select popular targets. For each forgery $i$, $\mathrm{target}_i$ is the targeted identity. We define a popularity function $\mathrm{Pop}(\cdot)$ as the total number of times a target is manipulated in the entire dataset. The conformity is the average popularity of an agent's chosen targets:
    \begin{equation}
        T_k^{\mathrm{conf}} = \frac{1}{|\mathrm{forgeries}_k|} \sum_{i \in \mathrm{forgeries}_k} \mathrm{Pop}(\mathrm{target}_i).
    \end{equation}
\end{itemize}
These three metrics form the numerical vector $\mathbf{v}_k$. For the qualitative component $\mathbf{c}_k$, we prompt GPT-4V to generate a stylistic preference description by analyzing a sample of $L$ forgeries from a creator's work, where $L$ is a predefined hyperparameter. Together, they form the agent's complete ``forgery gene,'' $\mathbf{p}_k = (\mathbf{v}_k, \mathbf{c}_k)$, which drives its decision-making.

\paragraph{Memory Module.} To enable continuous refinement, the memory module maintains two memory types: \textit{factual memory} for objective details of past edits, and \textit{evaluative memory} for subjective assessments (e.g., seam visibility). Memories are logged in structured JSON, containing not only operational data but also cognitive states like high-level plans. This supports three key operations: memory retrieval, writing, and an LLM-driven reflection process to analyze outcomes and guide future actions.

\paragraph{Action Module.} The Action Module translates intent into action. An agent's action at time $t$, denoted $\mathrm{Action}_k^{(t)}$, is a pair $(\mathrm{Edit}(\cdot), \mathrm{Desc}(\cdot))$ consisting of a visual edit and a textual description. The visual edit is a sequential application of operators:
\begin{equation}
    \mathrm{Edit}(\mathbf{x}; \mathbf{p}_k, \mathcal{M}_k) = O_n(\dots O_1(\mathbf{x}; \theta_1)\dots; \theta_n),
\end{equation}
where each operator $O_i$ is chosen from a toolbox $\mathcal{O}_{ops}$ (including methods like Flux Pro and Deepfake APIs), and its parameters $\theta_i$ are determined by the agent's profile $\mathbf{p}_k$ and its memory $\mathcal{M}_k$. The textual description $\mathrm{Desc}(\cdot)$ can be either an accurate caption or an intentionally misleading statement.

\paragraph{Adaptive Rejection Sampling (ARS).} The agents leverage their cognitive architecture within an iterative algorithm to synthesize and curate the final blueprints. To ensure the generated data is both diverse and challenging, we use an ARS mechanism as a quality control gate. A candidate blueprint $(\mathbf{x}'_i, \mathbf{c}'_i)$ is scored using a fused metric $s_i$:
\begin{equation}
    s_i = \lambda \, s_i^\mathrm{LLM} + (1 - \lambda) \, s_i^\mathrm{disc},
\end{equation}
where $s_i^\mathrm{disc}$ is the score from an external forgery detector, $s_i^\mathrm{LLM}$ is the agent's internal quality assessment based on its memory, and $\lambda$ is a weighting hyperparameter.A sample is accepted if its score $s_i$ exceeds an adaptive threshold $\tau$. To ensure the mechanism is logically sound and robust, the threshold determination includes an initial warm-up phase. For the first $N_warmup$ samples, we employ a fixed, lenient threshold, $\tau_warmup$, to build a diverse initial pool of accepted samples. After the warm-up phase, the threshold $\tau$ becomes fully data-driven and is periodically updated to the q-th quantile of the scores of all previously accepted samples:
\begin{equation}
    \tau = \text{Quantile}(\{s_j \mid j \in \text{Accepted Samples}\}, q),
\end{equation}
where $\{s_j\}$ is the set of scores from all accepted samples and $q \in [0, 1]$ is the single hyperparameter defining the rejection rate.

% A sample is accepted if $s_i$ exceeds an adaptive threshold $\tau$, which tightens over time to progressively favor higher-quality examples:
% \begin{equation}
%     \tau = \tau_0 + \omega \cdot N_\mathrm{accepted},
% \end{equation}
% where $\tau_0$ is the initial minimum threshold, $\omega$ is a tightening-rate parameter, and $N_\mathrm{accepted}$ is the count of accepted samples.

 \vspace{0.2em}
\noindent \textbf{Data Generation Pipeline.} The generation process unfolds in five structured steps:
\begin{enumerate}
 \item \textbf{Multi-round Forgery.} 
Given a set of real face images $\mathcal{X}_\mathrm{real}$, the agent's \textit{Action Module} generates a candidate by first formulating and then executing a multi-step forgery plan. For each random selected image $\mathbf{x} \in \mathcal{X}_\mathrm{real}$, the agent constructs a sequence of operators $(\mathcal{O}_1, \dots, \mathcal{O}_n)$ by sampling from a comprehensive toolbox $\mathcal{O}_\mathrm{ops}$. This toolbox is populated with diverse forgery instruments, categorized into:
\begin{itemize}
    \item \textbf{Identity Manipulation:} Face-swapping methods (e.g., based on DeepFaceLab, FaceSwap).
    \item \textbf{Attribute \& Expression Editing:} GAN-based editors for expressions or attributes like age and gender (e.g., StarGAN, AttGAN).
    \item \textbf{Style-Based Synthesis:} GAN-inversion techniques that allow for fine-grained stylistic edits (e.g., SBI).
    \end{itemize}
The selection of this operator chain is a probabilistic process, directly governed by the agent's \textit{Profile Module} ($\mathbf{p}_k$). The agent maintains a probability distribution over the tools in $\mathcal{O}_\mathrm{ops}$, where the likelihood of choosing a specific tool is weighted by its stylistic taste ($\mathbf{c}_k$) and methodological diversity trait ($T_k^{\mathrm{div}}$). For instance, an agent profiled for high-realism forgeries might have a higher probability of first selecting a face-swap operator, and then selecting a blending operator as a second step.

The execution of this plan, $\mathbf{x}'_i = \mathcal{O}_n(\dots \mathcal{O}_1(\mathbf{x})\dots)$, produces the final forged image. Following this, a textual corresponding forgery description $\mathbf{c}'_i = \mathrm{Desc}(\cdot)$ is generated, and the agent assigns the image-text consistency label $\delta'_i \in \{0, 1\}$ based on its strategic intent.

\item \textbf{Scoring and Filtering.} 
Each candidate is evaluated by two models pre-trained on the FF++ dataset. Xception and LLaVA, computes the agent's internal quality and consistency score $s_i^\mathrm{LLM} = f_\mathrm{agent}(\mathbf{x}'_i, \mathbf{c}'_i, \mathcal{M}_k)$. These scores are fused into a single metric $s_i$.Determine whether they are challenge samples based on the confidence score and the probability value of the LLaVA text output.

\item \textbf{Memory Update.} 
Both successful and rejected forgery attempts are logged into the \textit{Memory Module}. After a fixed number of iterations, the agent initiates a reflection phase, analyzing past outcomes to refine its forgery techniques, e.g., blending or style adjustments, for subsequent rounds.

\item \textbf{Output Data.} 
The retained forged samples, denoted as $\{(\mathbf{x}'_i, \mathbf{c}'_i, y_i=1, \delta'_i)\}$, are combined with the real face samples from $\mathcal{X}_\mathrm{real}$, which are formatted as $\{(\mathbf{x}_j, \mathbf{c}_j, y_j=0, \delta_j=1)\}$, to form the final multimodal dataset $\mathcal{D}$.

\end{enumerate}

 \begin{table*}[t]
\vspace{-1.2em}
    \renewcommand\arraystretch{1.1}
    \centering
    \caption{\textbf{Frame-level} cross-database evaluation from FF++(HQ) to DFD, DFDC-P, Wild Deepfake, and Celeb-DF in terms of AUC (\%) and EER (\%). The FF++ results represent intra-domain performance, while others represent generalization to unseen domains. The \textbf{best} results are indicated in bold, and the \underline{second-best} results are underlined.}
    \vspace{-0.8em}
    \label{table:1}
    \resizebox{0.7\textwidth}{!}{%
    \begin{tabular}{l|cc|cccccccc}
    \toprule
    \multirow{2}*{Method} & \multicolumn{2}{c|}{\textit{FF++}}&\multicolumn{2}{c}{DFD} & \multicolumn{2}{c}{DFDC-P} & \multicolumn{2}{c}{Wild Deepfake} &\multicolumn{2}{c}{Celeb-DF}\\
    \cmidrule(lr){2-3} \cmidrule(lr){4-5} \cmidrule(lr){6-7} \cmidrule(lr){8-9} \cmidrule(lr){10-11}
    & AUC& EER &AUC& EER& AUC& EER& AUC& EER& AUC& EER\\
    \midrule
    Xception~\cite{chollet2017xception}& 99.09&3.77 &87.86& 21.04& 69.80& 35.41&66.17 &40.14 & 65.27& 38.77\\
    EN-b4~\cite{tan2019efficientnet}    &99.22 &3.36& 87.37        & 21.99        & 70.12        & 34.54        & 61.04                 & 45.34                 & 68.52           & 35.61         \\
    Face X-ray~\cite{tan2019efficientnet}& 87.40&-& 85.60&-& 70.00&      -          &       -                 &       -                 & 74.20                 &       -         \\
    F3-Net~\cite{qian2020thinking}       & 98.10& 3.58 & 86.10      & 26.17      &    72.88      &    33.38     & 67.71                   &    40.17                & 71.21         &    34.03          \\
    MAT~\cite{multitask} & 99.27&3.35& 87.58        & 21.73        & 67.34        & 38.31        & 70.15                   & 36.53                   & 70.65           & 35.83         \\
    GFF~\cite{luo2021generalizing}         & 98.36& 3.85 & 85.51        & 25.64        & 71.58        & 34.77        & 66.51                   & 41.52                   & 75.31           & 32.48         \\
    LTW~\cite{sun2021domain}         & 99.17& 3.32  & 88.56        & 20.57        & 74.58        & 33.81        & 67.12                   & 39.22                   & 77.14           & 29.34          \\
    LRL~\cite{chen2021local}   & \underline{99.46}&\underline{3.01}& 89.24        & 20.32        & 76.53        & 32.41        & 68.76                   & 37.50                   & 78.26           & 29.67         \\
    DCL~\cite{sun2021dual}         & 99.30&3.26& 91.66        & 16.63        & \underline{76.71}        & 31.97        & \underline{71.14}                   & \underline{36.17}                   & 82.30           & 26.53    \\
    PCL+I2G~\cite{zhao2021learning}& 99.11&-& -        & -        & -        & -        & -         & -                   & 81.80           & -    \\
    SBI~\cite{shiohara2022detecting}     & 88.33&20.47&88.13&17.25&76.53&\underline{30.22}&68.22&38.11&80.76&26.97 \\
    UIA-ViT~\cite{zhuang2022uia} &-&-&94.68&-&75.80&-&-&-&\underline{82.41}&- \\
    RECCE~\cite{cao2022end} &99.32&3.38&89.91&19.95&75.88&32.41&67.93&39.82&70.50&35.34 \\
    UCF~\cite{yan2023ucf} &97.05&-&80.74&-&75.94&-&-&-&75.27&- \\
    FFTG~\cite{sun2025towards}         & 99.23&3.12& \textbf{94.79}      & \underline{15.31}      & \underline{84.74}      & \underline{23.43}      & \underline{83.55}               & \underline{24.40}               & \underline{84.80}           & \underline{22.73}         \\
    \midrule
     \textbf{Ours}         & \textbf{99.50}&\textbf{2.97}& \underline{93.25}      & \textbf{13.04}      & \textbf{88.10}      & \textbf{19.19}      & \textbf{86.50}                  & \textbf{21.87}                  & \textbf{87.10}              & \textbf{20.12}         \\
    \bottomrule
    \end{tabular}%
    }
    \vspace{-0.6em}
\end{table*}

 \vspace{-0.5em}
 \subsection{Phase 2: Social Interaction Trajectory Collection}

\paragraph{Positive-Negative Sample Construction. }Traditional approaches to data construction for forgery detection frequently depend on binary ``real'' or ``fake'' labels assigned at the image level, often disregarding the pivotal role of social discourse and user reactions in the dissemination of deepfakes. To address this shortcoming, we propose an innovative framework that simulates multi-user interactions within a social media environment, thereby aligning with the realistic dynamics of the ``social interaction--deepfake'' interplay. 

\noindent \textbf{Socal Intent Simulation. }Drawing inspiration from observable behaviors on social media platforms, we introduce five distinct user roles, each powered by MLLMs. These roles engage with forged images through a variety of actions, such as viewing, commenting, sharing, and labeling, thereby emulating a wide range of user interactions:

\begin{itemize}
    \item \textbf{Watcher}: Frequently designates content as ``liked'' or ``interesting'' but rarely investigates its authenticity.
    \item \textbf{Explorer}: Compares multiple posts related to the same event, enhancing the probability of detecting forgery artifacts through comparative analysis.
    \item \textbf{Critic}: Emphasizes quality and credibility, often highlighting suspicious forgeries in comments or reports.
    \item \textbf{Chatter}: Susceptible to misinformation due to social influences, yet capable of correction through group discussions.
    \item \textbf{Poster}: Reposts or re-edits content, amplifying the propagation of forged images across platforms.
\end{itemize}

Collectively, these roles simulate a rich and varied set of user interactions, yielding textual responses that closely resemble those observed in authentic social media contexts.

 \vspace{0.2em}
\noindent \textbf{Hard Negative Generation : }To amplify the complexity of text-image inconsistencies, we introduce the \textit{Gemini Auditor}, a specialized role engineered to produce intentionally deceptive statements. For example, the Gemini auditor might designate an evidently spliced image as ``100\% authentic'' or manipulate attributes such as gender or identity, thereby injecting ambiguity into the deepfake scenario. These adversarial assertions facilitate the creation of robust negative samples by inducing pronounced conflicts between text and image content, compelling detection models to confront deceptive pairings and bolstering their resilience.

 \vspace{0.2em}
\noindent \textbf{Social Environment Labeling. }The interactions among the Watcher, Explorer, Critic, Chatter, Poster, and Gemini roles generate a diverse collection of comments, labels, and edits for each forged image. When integrated with ground-truth labels (e.g., real or fake), these interactions enable the automated construction of positive and negative sample pairs based on the consistency between text and image content. Negative samples arise in cases such as a forged image ($y = 1$) paired with a claim asserting it is ``perfectly real,'' or a real image ($y = 0$) accompanied by a declaration of ``obvious forgery.'' Conversely, positive samples emerge when the text accurately reflects the image’s authenticity or when social feedback, such as a Critic identifying artifacts, rectifies an initial mislabeling.We formalize this labeling process with the function $\delta(x', c')$:
 \begin{equation}
    \begin{cases}
1, & \text{if } y = 1 \text{ and } c' \text{ claims ``perfectly real''}, \\
1, & \text{if } y = 0 \text{ and } c' \text{ claims ``obvious forgery''}, \\
0, & \text{otherwise},
\end{cases}
\end{equation}
\noindent where $\delta = 1$ denotes a negative sample indicative of a text--image mismatch, and $\delta = 0$ signifies alignment or a corrected positive sample. This methodology capitalizes on social interactions to produce nuanced, context-sensitive labels that mirror the intricacies of real-world scenarios.
%================================================ TABLE 1 & 2 (SOTA Comparison) ================================================

\begin{table*}[!h]
%\vspace{-1.2em}
\renewcommand\arraystretch{1.1}
\centering
\caption{Assessing detector robustness to diverse manipulation algorithms within the FF++ dataset. We report frame-level AUC (\%) against six techniques specified in DF40 \citep{yan2024df40}.}
\vspace{-0.8em}
\label{tab:tab2}
\resizebox{0.85\textwidth}{!}{%
\begin{tabular}{l c c c c c c c c}
\toprule
\textbf{Method} & \textbf{Venue} & \textbf{uniface} & \textbf{e4s}& \textbf{facedancer} & \textbf{fsgan} & \textbf{inswap} & \textbf{simswap} & \textbf{Avg.} \\
\midrule
RECCE \citep{cao2022end} & CVPR 2022 & 84.2 & 65.2 & 78.3 & 88.4 & 79.5 & 73.0 & 78.1 \\
SBI \citep{shiohara2022detecting} & CVPR 2022 & 64.4 & 69.0 & 44.7 & 87.9 & 63.3 & 56.8 & 64.4 \\
CORE \citep{ni2022coreconsistentrepresentationlearning} & CVPRW 2022 & 81.7 & 63.4 & 71.7 & \underline{91.1} & 79.4 & 69.3 & 76.1 \\
IID \citep{huang2023implicit} & CVPR 2023 & 79.5 & \underline{71.0} & 79.0 & 86.4 & 74.4 & 64.0 & 75.7 \\
UCF \citep{yan2023ucf} & ICCV 2023 & 78.7 & 69.2 & \underline{80.0} & 88.1 & 76.8 & 64.9 & 76.3 \\
LSDA \citep{yan2023transcending} & CVPR 2024 & \underline{85.4} & 68.4 & 75.9 & 83.2 & \underline{81.0} & 72.7 & 77.8 \\
CDFA \citep{lin2024CDFA} & ECCV 2024 & 76.5 & 67.4 & 75.4 & 84.8 & 72.0 & 76.1 & 75.4 \\
ProgressiveDet \citep{cheng2024can} & NeurIPS 2024 & 84.5 & \underline{71.0} & 73.6 & 86.5 & 78.8 & \underline{77.8} & \underline{78.7} \\
\midrule
\textbf{Ours} & - & \textbf{96.3} & \textbf{92.4} & \textbf{92.9} & \textbf{94.8} & \textbf{92.4} & \textbf{94.6} & \textbf{93.9} \\
\bottomrule
\end{tabular}%
}
\vspace{-1.0em}
\end{table*}

%================================================ TABLE 3,4,5,6,7 (Ablation) ================================================
\begin{table*}[t]
%\vspace{-0.8em}
    \renewcommand\arraystretch{1.1}
    \centering
    \caption{Comparison of different annotation approaches. We report precision, recall and F1-score for annotation quality evaluation, AUC (\%) and EER for CLIP-based forgery detection and ACC (\%) and explanation quality (Precision/Recall) for MLLMs evaluation on FF++ and Celeb-DF (CDF) datasets.}
    \vspace{-0.8em}
    \label{table:3}
    \resizebox{0.83\textwidth}{!}{%
    \begin{tabular}{l|ccc|cc|cccc}
    \toprule
    \multirow{2}*{Method} & \multicolumn{3}{c|}{{Annotation Evaluation}}& \multicolumn{2}{c|}{CLIP Evaluation} & \multicolumn{4}{c}{MLLM Evaluation} \\
    \cmidrule{2-10}
    & Precision& Recall &F1& AVG-AUC& AVG-EER&  FF++-ACC&CDF-ACC&Precision& Recall \\
    \hline
    w/o Text&-&-&-&84.36&20.64&50.13&65.30&10.41&8.10\\
    DD-VQA (Human)&62.46&51.52&52.06&88.25&18.04&73.54&65.60&62.94&53.62\\
    GPT-4o-mini&61.27&44.00&47.18&87.56&19.21&94.84&73.98&58.26&41.85\\
     \textbf{Ours}&\textbf{94.41}&\textbf{60.04}&\textbf{69.06}&\textbf{91.23}&\textbf{16.35}&\textbf{96.35}&\textbf{77.98}&\textbf{89.02}&\textbf{59.02}\\
    \bottomrule
    \end{tabular}
    }
    \vspace{-1.0em}
\end{table*}

\begin{table}[t]
\centering
\caption{Ablation on the sequential training strategy. Models are first pre-trained on a base dataset and then fine-tuned with data generated by Agent4FaceForgery (A4FF). The results demonstrate significant performance improvements across multiple backbone models, confirming the effectiveness of our generated data for augmentation.}
\vspace{-0.8em}
\label{tab:sequential_training_ablation}
\resizebox{0.7\linewidth}{!}{%
\begin{tabular}{lccc}
\toprule
\textbf{Model} & \textbf{A4FF Data} & \textbf{AUC(\%) $\uparrow$} & \textbf{EER(\%) $\downarrow$} \\
\midrule
\multirow{2}{*}{Phi-3.5}     & $\times$     & 81.5          & 25.3          \\
                             & \checkmark   & \textbf{90.4} & \textbf{19.0} \\
\midrule
\multirow{2}{*}{Qwen-VL 2.5} & $\times$     & 82.7          & 26.1          \\
                             & \checkmark   & \textbf{91.7} & \textbf{18.4} \\
\midrule
\multirow{2}{*}{LLaVA}       & $\times$     & 83.2          & 24.8          \\
                             & \checkmark   & \textbf{92.2} & \textbf{16.8} \\
\bottomrule
\end{tabular}%
}
\vspace{-0.6em}
\end{table}

\begin{table}[t]
  \centering
  \caption{Results of different backbone models with and without Agent4FaceForgery on different datasets.}
  \vspace{-0.8em}
  \label{tab:model}
  \resizebox{0.9\linewidth}{!}{
  \begin{tabular}{lcc|cc|cc}
    \toprule
    \multirow{2}{*}{Backbone} 
      & \multirow{2}{*}{CLIP} 
      & \multirow{2}{*}{Agent} 
      & \multicolumn{2}{c}{WDF} 
      & \multicolumn{2}{c}{DFDC-P} \\
    \cmidrule(lr){4-5} \cmidrule(lr){6-7}
      & & 
      & AUC & EER 
      & AUC & EER \\
    \midrule
    \multirow{2}{*}{Xception~\cite{xception}} 
      & $\times$ & $\times$ 
      & 66.17 & 40.14 
      & 69.80 & 35.41 \\
      & $\times$ & \checkmark 
      & \textbf{73.14} & \textbf{35.12} 
      & \textbf{77.64} & \textbf{28.09} \\
    \midrule
    \multirow{2}{*}{EN-B4~\cite{efficientnet}} 
      & $\times$ & $\times$ 
      & 61.04 & 45.34 
      & 70.12 & 34.54 \\
      & $\times$ & \checkmark 
      & \textbf{73.78} & \textbf{34.57} 
      & \textbf{80.04} & \textbf{27.62} \\
    \midrule
    \multirow{3}{*}{ViT-L~\cite{clip}} 
      & $\times$ & $\times$ 
      & 65.39 & 38.11 
      & 68.03 & 36.06 \\
      & $\times$ & \checkmark 
      & 76.12 & 30.07 
      & 77.54 & 28.71 \\
      & \checkmark & \checkmark 
      & \textbf{79.32} & \textbf{26.51} 
      & \textbf{79.44} & \textbf{28.79} \\
    \midrule
    \multirow{3}{*}{ViT-B~\cite{clip}} 
      & $\times$ & $\times$ 
      & 69.92 & 36.20 
      & 71.33 & 33.76 \\
      & $\times$ & \checkmark 
      & 73.55 & 30.12 
      & 81.79 & 26.31 \\
      & \checkmark & \checkmark 
      & \textbf{86.50} & \textbf{21.87} 
      & \textbf{88.10} & \textbf{19.19} \\
    \bottomrule
  \end{tabular}}
  \vspace{-1.6em}
\end{table}

\begin{table}[t]
\centering
\vspace{-0.3em}
\caption{Ablation study regarding the effectiveness of each proposed module via cross-dataset evaluation. The results show an incremental benefit in each module.}
\vspace{-0.8em}
 \label{tab:module}
 \resizebox{\linewidth}{!}{
\begin{tabular}{@{}ccccccc@{}}
\toprule
\multirow{2}{*}{\textbf{Method}} & \multicolumn{3}{c}{\textbf{Modules}} & \multicolumn{3}{c}{\textbf{Metrics (AUC(\%) \textbar{} AP(\%) \textbar{} EER(\%))}} \\ \cmidrule(lr){2-4} \cmidrule(lr){5-7}
 & \textbf{FT} & \textbf{ARS} & \textbf{PNS} & \textbf{CDF} & \textbf{DFD} & \textbf{DFDC} \\ \midrule
\makecell[c]{LLaVA} & - & - & - & \metrics{51.8}{68.0}{49.3} & \metrics{69.3}{94.9}{37.2} & \metrics{57.4}{60.4}{45.3} \\
Only FT & $\checkmark$ & - & - & \metrics{83.2}{90.8}{24.8} & \metrics{91.5}{98.6}{16.3} & \metrics{82.5}{90.5}{22.1} \\
Only ARS & - & $\checkmark$ & - & \metrics{88.0}{93.7}{21.0} & \metrics{92.1}{98.8}{16.8} & \metrics{84.2}{86.5}{22.3} \\
Only PNS & - & - & $\checkmark$ & \metrics{91.0}{95.0}{17.5} & \metrics{93.8}{99.0}{16.2} & \metrics{85.5}{87.2}{20.5} \\
\midrule
 \textbf{Ours} & \textbf{$\checkmark$} & \textbf{$\checkmark$} & \textbf{$\checkmark$} & \metrics{92.2}{95.6}{16.8} & \metrics{94.9}{99.2}{15.7} & \metrics{86.7}{88.0}{19.5} \\
 \bottomrule
\end{tabular}}%
\vspace{-0.5em}
\end{table}

\begin{table}[t]
\centering
\caption{Ablation on the number of agents in the social simulation. Performance is evaluated on the DFD and Celeb-DF datasets. The results show diminishing returns as the number of agents increases, with 12 agents providing only marginal gains over 6, which justifies our agent count configuration and balances performance with computational cost.}
\vspace{-0.8em}
\label{tab:agent_number_ablation}
\resizebox{\linewidth}{!}{%
\begin{tabular}{lccccc}
\toprule
\multirow{2}{*}{\textbf{Configuration}} & \multicolumn{2}{c}{\textbf{DFD}} & \multicolumn{2}{c}{\textbf{Celeb-DF}} & \multirow{2}{*}{\textbf{Time(h)}} \\
\cmidrule(lr){2-3} \cmidrule(lr){4-5}
 & AUC(\%) $\uparrow$ & EER(\%) $\downarrow$ & AUC(\%) $\uparrow$ & EER(\%) $\downarrow$ & \\
\midrule
Baseline (No Social Sim) & 88.1 & 20.0 & 74.5 & 31.0 & 3.8 \\
2 Agents & 89.8 & 18.1 & 77.9 & 27.8 & 4.5 \\
4 Agents & 91.3 & 16.5 & 81.5 & 24.6 & 5.3 \\
6 Agents & 92.8 & 14.9 & 85.3 & 22.0 & 6.1 \\
12 Agents & \textbf{93.0} & \textbf{14.5} & \textbf{85.8} & \textbf{21.6} & 7.5 \\
\bottomrule
\end{tabular}%
}
\vspace{-1.5em}
\end{table}

\vspace{-0.6em}
\section{Experiments}
\subsection{Experimental Settings}
\noindent \textbf{Datasets.} 
Following standard practice~\cite{f3net,lsda,recce}, we built our model based on FF++~\cite{ff++}. We evaluate performance on several benchmarks, including Celeb-DF (V2)~\cite{celebdf}, DFDC~\cite{dfdc}, WildDeepfake~\cite{wildfake}, DFD~\cite{dfdc}, and DFDC-P~\cite{dfdcp}. Robustness is further assessed using the DF40 protocol~\cite{lsda,vlffd2025}.

\vspace{0.3em}
\noindent \textbf{Implementation Details.}
We use LLava~\cite{llava-1.5}, as the backbone model, and RetinaFace~\cite{retinaface} to detect facial areas and scaled the face image to $224\times224$ with a patch size of $16$. We trained the model using the Adam optimizer with the learning rate set to 3e-6. The frame setting is consistent with common practices in~\cite{recce,f3net,yan2023ucf}. \textit{In our experiments, we generated approximately 25k image-text pairs.}

\vspace{-0.4em}
\subsection{Comparison with State-of-the-Art Methods}
\noindent \textbf{Cross-Dataset Generalization.}
We first evaluated the detector's ability to generalize to completely unseen datasets. As shown in Table~\ref{table:1}, all models were trained on FF++ (HQ) and tested on datasets including DFD, DFDC-P, WildDeepfake, and Celeb-DF. Our model achieved top-tier or second-best performance across all cross-dataset scenarios. For instance, our model reached an AUC of 87.10\% on highly challenging Celeb-DF, and it achieved an AUC of 86.50\% on WildDeepfake. This superior generalization performance stems from the higher ecological validity of our simulated data. By modeling diverse human intents through the Profile module and complex social interactions via the PNS module, our data better captures the fundamental characteristics of real-world forgeries, allowing the model to generalize beyond overfitting to specific dataset artifacts.

\vspace{0.2em}
\noindent \textbf{Robustness to Diverse Manipulations.}
Next, we assessed the model's robustness against a variety of unseen forgery algorithms using the DF40 protocol. As shown in Table~\ref{tab:tab2}, our model achieved an average AUC of 93.9\% when tested against six advanced forgery techniques like uniface, e4s, and simswap, significantly outperforming all existing SOTA methods. This exceptional robustness further demonstrates the value of our data generation framework. Through the ARS mechanism that filters for high-difficulty samples and the diverse agent Profiles, our framework generates training data that covers a broader spectrum of forgery traces. This process enables the detector to learn more essential and universal forgery features, rather than simply memorizing the patterns of a few specific known forgery types.

\vspace{-0.6em}
\subsection{Ablation Study}
\noindent \textbf{Effectiveness of Agent-Generated Annotations.}
A core claim of our work is that simulating the forgery lifecycle yields higher-quality multimodal annotations. We validated this through a multi-faceted comparison of data from Agent4FaceForgery against human annotations (DD-VQA) and direct MLLM annotations (GPT-4o). As shown in Table~\ref{table:3}, our approach excels on multiple fronts. First, in terms of annotation quality, our data achieved a Precision of 94.41\% and an F1-score of 69.06\%, far surpassing other methods. This demonstrates that our agent simulation (especially the `Memory` and `PNS` modules) generates text that is better aligned with visual evidence and less prone to hallucination. Second, on downstream tasks, models trained with our data performed best, whether training a standard CLIP model (achieving an AVG-AUC of 91.23\%) or fine-tuning an MLLM (LLaVA), which reached an accuracy of 77.98\% on Celeb-DF. These combined results confirm that our simulation process produces superior multimodal training data, leading to tangible improvements in detector performance, generalization, and interpretability.

\vspace{0.3em}
\noindent \textbf{Sequential Training with A4FF Data.}
Agent4FaceForgery is designed as a data generation and augmentation framework. Our standard training strategy is sequential: pre-training a model on a base dataset (FF++) and then fine-tuning it with our generated data to enhance performance. The effectiveness of this strategy is clearly demonstrated in Table~\ref{tab:sequential_training_ablation}. After augmentation with Agent4FaceForgery (A4FF) data, the performance of various models, including Phi-3.5, Qwen-VL 2.5, and LLaVA, improved significantly. For instance, the LLaVA model's AUC improved from 83.2\% to 92.2\%, and its EER dropped from 24.8\% to 16.8\%. This strongly validates the value of our framework as a powerful data augmentation tool.

\vspace{0.3em}
\noindent \textbf{Impact on Different Backbone Architectures.}
To demonstrate the universal applicability of our generated data, we evaluated its effectiveness across four different backbone architectures: Xception, EfficientNet-B4, ViT-B, and ViT-L. The results in Table~\ref{tab:model} show that all architectures exhibited significant performance gains when trained with data from Agent4FaceForgery compared to using only standard FF++ data. It is particularly noteworthy that combining CLIP pre-training with our agent-generated data on a ViT-B backbone yielded an AUC of 86.50\% on WDF and 88.10\% on DFDC-P. This powerful synergy shows that our framework provides effective ecological signals for a wide range of models, greatly enhancing their generalization capabilities.

\vspace{0.3em}
\noindent \textbf{Effectiveness of Core Modules.}
We further conducted an ablation study to isolate the contribution of our framework's core components: Forgery Tree simulation (FT), ARS, and PNS. As detailed in Table~\ref{tab:module}, we started with a baseline LLaVA model trained only on FF++ (achieving just 51.8\% AUC on CDF). Adding data from the `FT` module alone boosts performance to 83.2\%, validating the benefit of simulating diverse forgery intents. The ARS and PNS modules also provided substantial gains individually, with the PNS module being particularly impactful (reaching 91.0\% AUC), highlighting the importance of simulating social context and generating challenging text-image consistency samples. The full system, with all modules working in concert, achieved the best performance (92.2\%AUC).  

\vspace{0.3em}
\noindent \textbf{Impact of Agent Number in Social Simulation.}
To justify our choice of agent count for the social simulation, we experimented with varying numbers of agents. Table~\ref{tab:agent_number_ablation} shows a clear trend of diminishing marginal returns. While performance on DFD and Celeb-DF consistently improves as the number of agents increases from 2 to 6, the gain from 6 to 12 agents is minimal. For instance, the Celeb-DF AUC increases from 85.3\% to only 85.8\%, at a significant additional time cost. Therefore, we selected 6 agents as our default configuration, striking a reasonable balance between performance and computational efficiency.

%\vspace{-1.0em}

 \begin{figure}[t] 
\vspace{-1.2em}
\centering 
\includegraphics[width=1.0\linewidth]{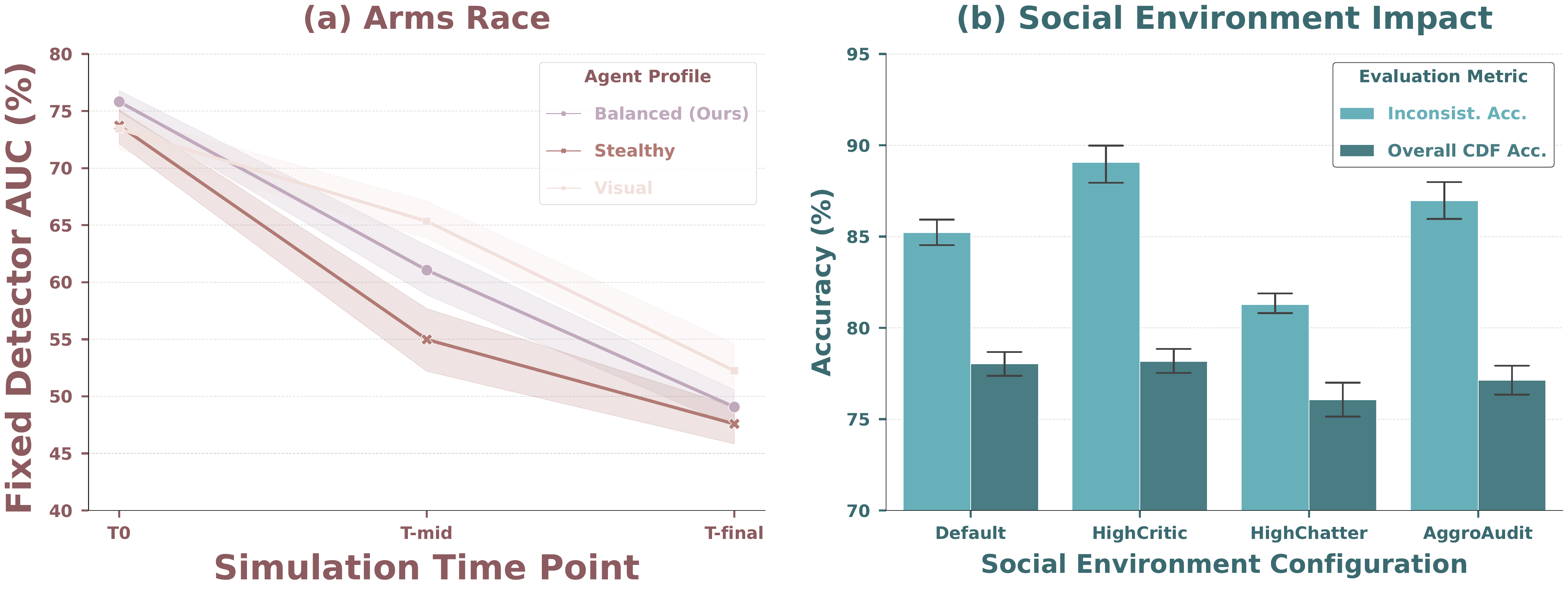}
\vspace{-1.8em}
\caption{Left (a): Comparison of forgery evasion capability (AUC, lower is better) evolution over simulation time for Agents with different profiles. Right (b): Comparison of MLLM detection accuracy (Inconsistency vs. Overall) when trained on data generated under different Social Environment configurations.}
\label{aba2} 
\vspace{-1.3em}
\end{figure}

%  \begin{figure}[t] 
% %\vspace{-1.1em}
% \centering 
% \includegraphics[width=1.0\linewidth]{Fig/simulation2.pdf}
% \vspace{-1.8em}
% \caption{Agent metrics in a filter bubble simulation comparing control (Ctrl) and treatment (Treat) groups. (a) Content diversity entropy over time. (b) Perceptual ability (recognition, resistance) at key epochs. (c) Judgment fixation (confidence, rebuttal frequency) over time.}
% \label{fig:filter_bubble_metrics}
% \label{aba3} 
% \vspace{-1.1em}
% \end{figure}

 % We also analyze the impact of core components within Agent4FaceForgery, focusing on agent profile influence on forgery evolution and social simulation effects on downstream detectors (Fig.~\ref{aba2}).

 \vspace{0.3em}
\noindent \textbf{Agent Profile Modulates Forgery Evolution.}
We investigate the impact of agent profiles on simulated forgery evolution against a fixed detector in Fig.~\ref{aba2}(a). Measuring evasion capability via AUC (lower is better), all profiles (Balanced, Stealthy, Visual) demonstrate adaptive generation, progressively reducing detectability from T0 to T-final, validating the efficacy of the Memory/Reflection and ARS modules. Crucially, the Stealthy profile achieves superior evasion, reaching the lowest final AUC ($\approx$48.5\%), while the 'Visual' profile lags. This confirms that the agent's profile, representing forgery intent, significantly modulates the adaptive generation process within the simulated arms race.

\vspace{0.3em}
\noindent \textbf{Social Simulation Enhances Detector Robustness.}
Fig.~\ref{aba2}(b) evaluates how the simulated social environment during data generation impacts the robustness of a trained MLLM detector. We assess MLLM accuracy on text-image inconsistency detection (Inconsist. Acc.) and overall generalization (Overall CDF Acc.). Training with data from the 'HighCritic' environment yields the most robust detector ($\approx$88.7\% Inconsist. Acc., $\approx$78.6\% CDF Acc.), demonstrating the value of incorporating simulated critical feedback. Conversely, the HighChatter environment degrades performance. The AggroAudit setting, featuring adversarial text generation, notably enhances robustness against inconsistencies ($\approx$87.1\% vs. $\approx$85.3\% baseline). These results underscore the importance of social simulation and context-aware PNS construction. By generating ecologically valid training data reflecting diverse social interactions and potential adversarial text, our framework demonstrably improves detector robustness to complex multimodal manipulations.

\begin{figure}[t] 
\vspace{-1.5em}
\centering 
\includegraphics[width=1.0\linewidth]{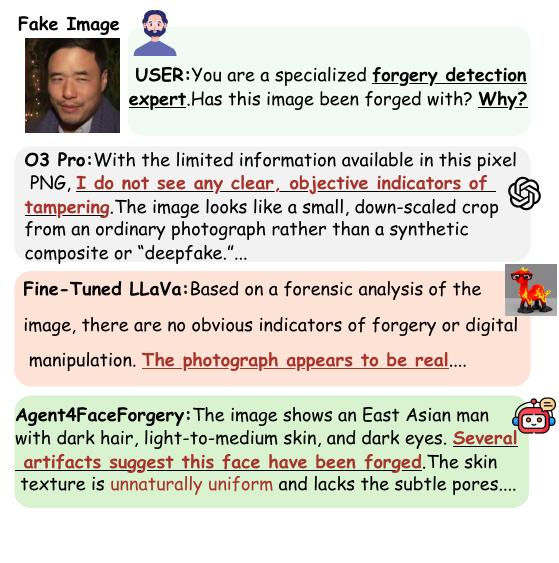}
\vspace{-4.2em}
\caption{ Qualitative examples in challenge scenarios.
} 
\label{example_agent} 
\vspace{-0.5em}
\end{figure}

\begin{figure}[t] 
\vspace{-0.3em}
\centering 
\includegraphics[width=0.8\linewidth]{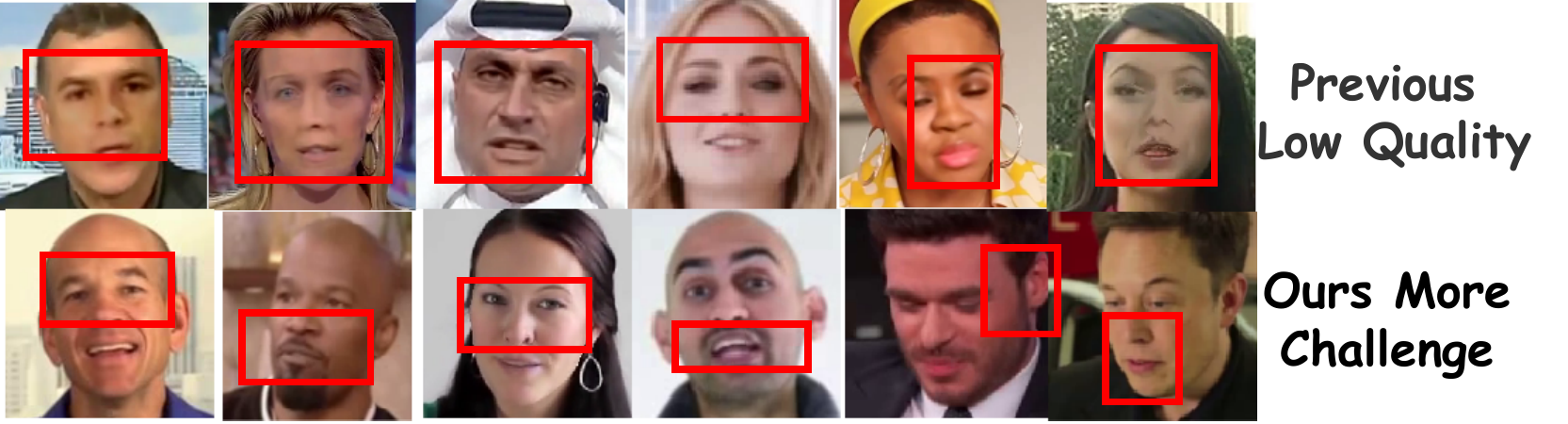}
%\vspace{-4.2em}
\caption{ Qualitative examples of Agent-Generated Images.
} 
\label{example_agent2} 
\vspace{-1.4em}
\end{figure}

\noindent \textbf{Qualitative Analysis.} As shown in Fig.~\ref{example_agent} showcases a qualitative analysis where our Agent4FaceForgery demonstrates superior forgery detection over other models. Presented with a forged image, baseline models O3 Pro and a fine-tuned LLaVA fail to identify any manipulation, stating a lack of "clear, objective indicators" and deeming the photograph "to be real." Conversely, Agent4FaceForgery accurately flags the image as forged, offering a succinct explanation: ``The skin texture is unnaturally uniform...". This highlights Agent4FaceForgery's enhanced ability to discern subtle forgery cues missed by other approaches. In addition, Fig.~\ref{example_agent2} provides qualitative examples of the high-fidelity and stylistically diverse portraits generated by our framework.

 \vspace{-0.7em}
\section{Conclusion}
We introduce Agent4FaceForgery, a multi-agent framework that simulates human behavior to generate realistic multimodal data for face forgery detection. Refined via adaptive rejection sampling, our synthesized data significantly boosts detector performance and generalization across challenging cross-domain benchmarks. The framework's extensible design offers insights into forgery tactics and supports future work in self-supervised learning to counter evolving threats.

\bibliography{aaai25}

\end{document}